\pdfoutput=1

\documentclass[11pt]{article}

\usepackage{emnlp2021}

\usepackage{times}
\usepackage{latexsym}
\usepackage{graphicx}
\usepackage{multirow}
\usepackage{adjustbox}
\usepackage{esvect}
\usepackage{booktabs}

\usepackage[T1]{fontenc}

\usepackage[utf8]{inputenc}

\usepackage{microtype}

\title{\textsc{ClauseRec:} A Clause Recommendation Framework for AI-aided Contract Authoring}

\author{Vinay Aggarwal \\
  Adobe Research \\
  \texttt{vinayagg@adobe.com} \\\And
  Aparna Garimella \\
  Adobe Research \\
  \texttt{garimell@adobe.com} \\ \And
  Balaji Vasan Srinivasan \\
  Adobe Research \\
  \texttt{balsrini@adobe.com} \\ \AND
  Anandhavelu N \\
  Adobe Research \\
  \texttt{anandvn@adobe.com} \\ \And
  Rajiv Jain \\
  Adobe Research \\
  \texttt{rajijain@adobe.com} \\
  }

\def\vinay#1{\textcolor{brown}{[Vinay: #1] }}
\def\aparna#1{\textcolor{red}{[Aparna: #1] }}
\def\bvs#1{\textcolor{blue}{[Balaji: #1] }}

\begin{document}
\maketitle
\begin{abstract}
Contracts are a common type of legal document that frequent in several day-to-day business workflows.
However, there has been very limited NLP research in processing such documents, and even lesser in generating them. 
These contracts are made up of clauses, and the unique nature of these clauses calls for specific methods to understand and generate such documents.
In this paper, we introduce the task of {\it clause recommendation}, as a first step to aid and accelerate the authoring of contract documents. 
We propose a two-staged pipeline to first predict if a specific clause type is relevant to be added in a contract, and then recommend the top clauses for the given type based on the contract context.
We pretrain BERT on an existing library of clauses with two additional tasks and use it for our prediction and recommendation.
We experiment with classification methods and similarity-based heuristics for clause relevance prediction, and generation-based methods for clause recommendation, and evaluate the results from various methods on several clause types. 
We provide analyses on the results, and further outline the advantages and limitations of the various methods for this line of research.


\end{abstract}

\section{Introduction}
A contract is a legal document between at least two parties that outlines the terms and conditions of the parties to an agreement.
Contracts are typically in textual format, thus providing a huge potential for NLP applications in the space of legal documents. 
However, unlike most natural language corpora that are typically used in NLP research, contract language is repetitive with high inter-sentence similarities and sentence matches \cite{simonson-etal-2019-extent},
calling for new methods specific to legal language to understand and generate contract documents.

A contract is essentially made up of {\it clauses}, which are provisions to address specific terms of the agreement, and which form the legal essence of the contract.
Drafting a contract involves selecting an appropriate template (with skeletal set of clauses), and customizing it for the specific purpose, typically via adding, removing, or modifying the various clauses in it.
Both these stages involve manual effort and domain knowledge, and hence can benefit from assistance from NLP methods that are trained on large collections of contract documents.
In this paper, we attempt to take the first step towards 
AI-assisted contract authoring, and introduce the task of {\it clause recommendation}, and propose a two-staged approach to solve it. 

\begin{figure*}[t]
    \centering
    \includegraphics[width=0.8\linewidth]{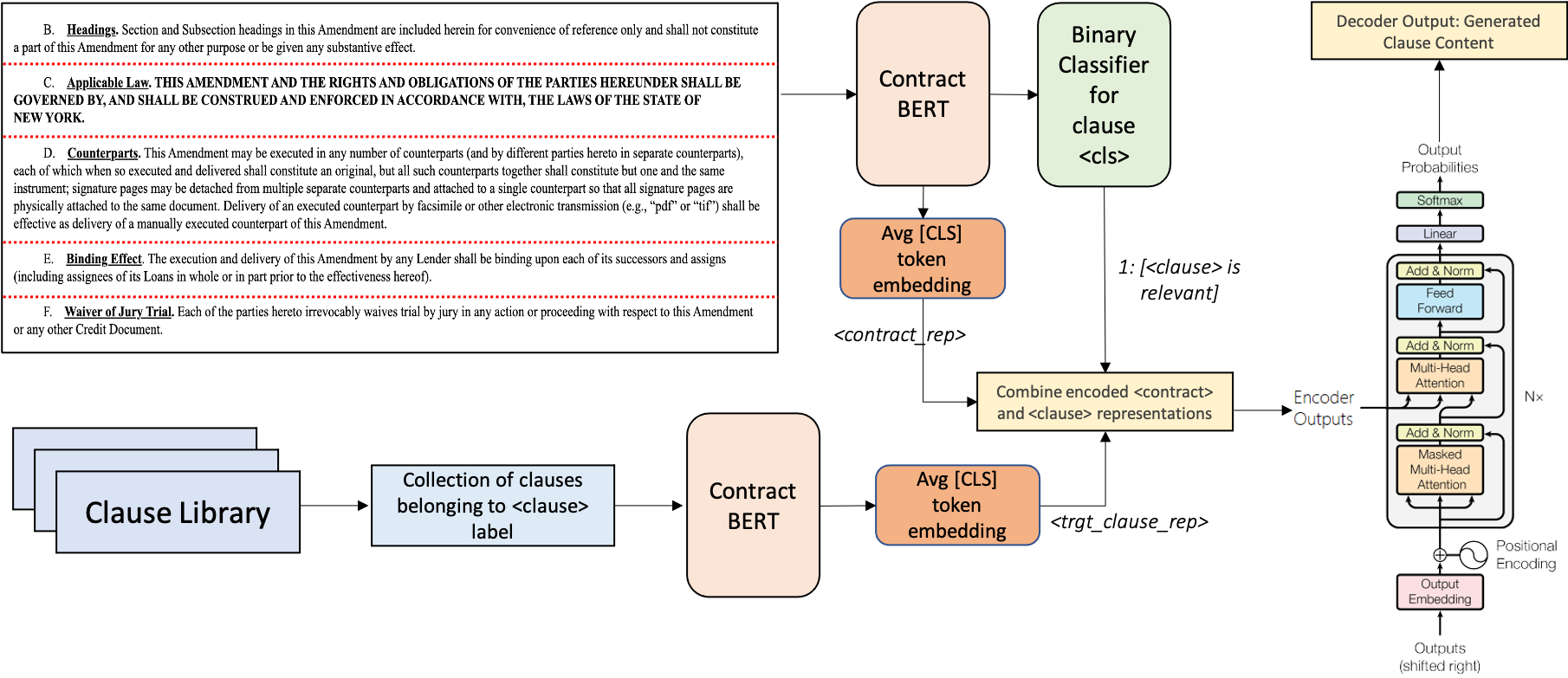}
    \captionsetup{justification=centering}
    \caption{{\sc ClauseRec} pipeline: Binary classification + generation for clause recommendation.}
    \label{fig:pipeline}
    \vspace{-0.15in}
\end{figure*}

There have been some recent works on item-based and content-based recommendations. \citet{wang-fu-2020-item} reformulated the next sentence prediction task in BERT \cite{devlin-etal-2019-bert} as next purchase prediction task to make a collaborative filtering based recommendation system for e-commerce setting. \citet{malkiel-etal-2020-recobert} introduced RecoBERT leveraging textual description of items such as titles to build an item-to-item recommendation system for wine and fashion domains. 
In the space of text-based content recommendations, \citet{bhagavatula-etal-2018-content} proposed a method to recommend citations in academic paper drafts without using metadata. 
However, legal documents remain unexplored, and it is not straightforward to extend these methods to recommend clauses in contracts, as these documents are heavily domain-specific and recommending content in them requires specific understanding of their language. 
In this paper, clause recommendation is defined as the process of automatically providing recommendations of clauses that may be added to a given contract while authoring it.
We propose a two-staged approach: first, we predict if a given {\it clause type} is relevant to be added to the given input contract; examples of clause types include {\it governing laws}, {\it confidentiality}, etc.
Next, if a given clause type is predicted as relevant, we provide context-aware recommendations of clauses belonging to the given type for the input contract.
We develop {\sc ContractBERT}, by further pre-training BERT using two additional tasks, and use it as the underlying language model in both the stages to adapt it to contracts.
To the best of our knowledge, this is the first effort towards developing AI assistants for authoring and generating long domain-specific legal contracts.
\section{Methodology}
A contract can be viewed as a collection of clauses with each clause comprising of: (a) the clause label that represents the type of the clause
and (b) the clause content. 
Our approach consists of two stages: {\bf (1)} {\it clause type relevance prediction}: predicting if a given clause type that is not present in the given contract may be relevant to it, and {\bf (2)} {\it clause recommendation}: recommending clauses corresponding to the given type that may be relevant to the contract.
Figure \ref{fig:pipeline} shows an overview of our proposed pipeline. 

\begin{figure}[t]
    \centering
    \includegraphics[width=0.45\textwidth]{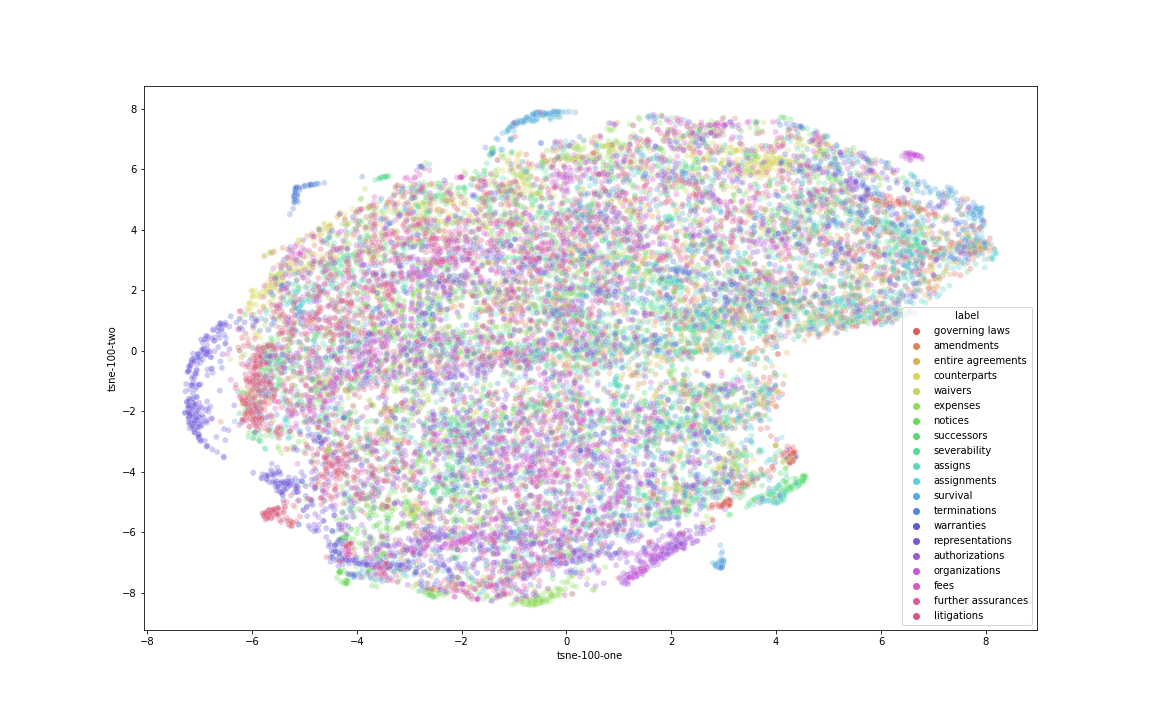}
    \captionsetup{justification=centering}
    \caption{Clustering of clauses using BERT Embedding}
    \label{fig:bert_tsne}
    \vspace{-0.15in}
\end{figure}

\begin{figure}[t]
    \centering
    \includegraphics[width=0.45\textwidth]{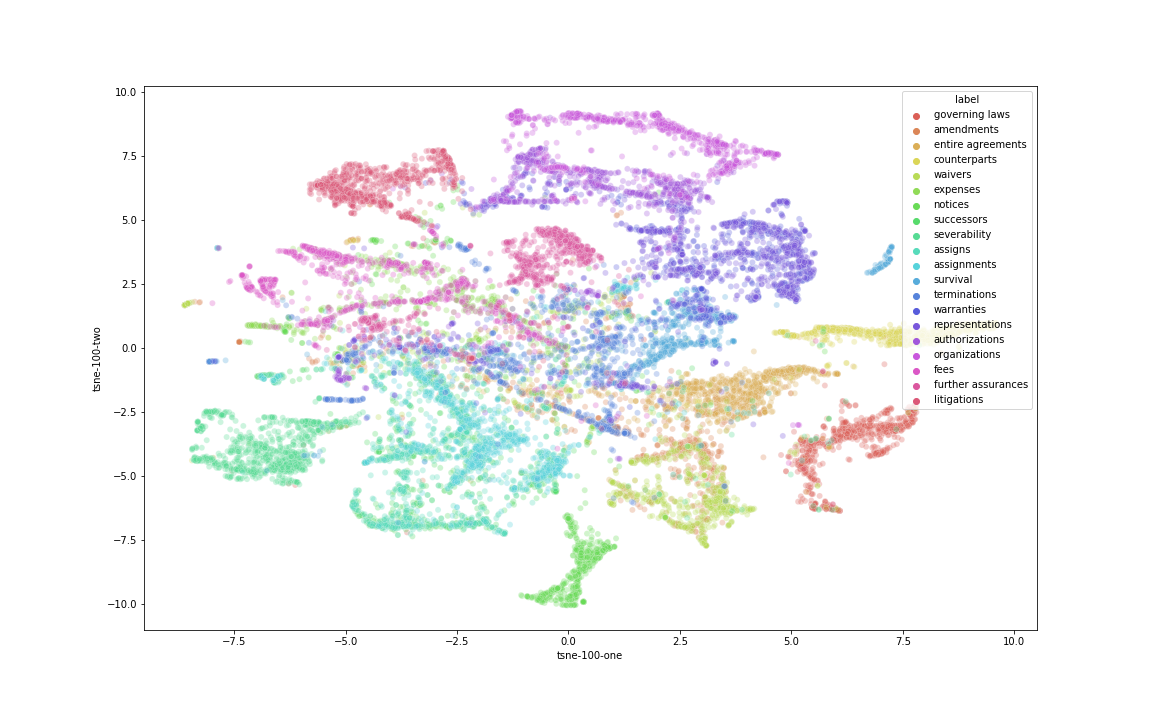}
    \captionsetup{justification=centering}
    \caption{Clustering of clauses using ContractBERT Embedding}
    \label{fig:cbert_tsne}
    \vspace{-0.15in}
\end{figure}

First, we build a model to effectively represent a contract by further pre-training BERT, a pre-trained Transformer-based encoder \cite{devlin-etal-2019-bert}, on contracts to bias it towards legal language.
We refer to the resulting model as \textbf{\textsc{ContractBERT}}.
In addition to masked language modelling and next sentence prediction, {\sc ContractBERT} is trained to predict {\it (i)} if the words in a clause label belong to a specific clause, and {\it (ii)} if two sentences belong to the same clause, enabling the embeddings of similar clauses to cluster together. Figure \ref{fig:bert_tsne} and \ref{fig:cbert_tsne} show the difference in the performance of BERT and {\sc ContractBERT} to get a meaningful clause embedding. BERT is unable to differentiate between the clauses of different types as it is unfamiliar with legal language. On the other hand, {\sc ContractBERT} is able to cluster similar clause types closely while ensuring the separation between clauses of two different types. 

\subsection{Clause Type Relevance Prediction}
Given a contract and a specific target clause type, the first stage involves predicting if the given type may be relevant to be added to the contract.
We train binary classifiers for relevance prediction for each of the target clause types. 
Given an input contract, we obtain its {\sc ContractBERT} representation as shown in Figure \ref{fig:pipeline}.
Since the number of tokens in the contracts are usually very large ($\gg512$), we obtain the contextual representations of each of the clauses present and average their [CLS] embeddings to obtain the contract representation \textit{ct\_rep}.
This representation is fed as input to a binary classifier which is a small fully-connected neural network that is trained using binary cross entropy loss.
We use a probability score of over $0.5$ as a positive prediction, i.e., the target clause type is relevant to the input contract. 

\subsection{Clause Content Recommendation}
Once a target clause type is predicted as relevant, the next stage is to recommend clause content corresponding to the given type for the contract.
We model this as a sequence-to-sequence generation task, where the the input includes the given contract and clause label, and the output contains relevant clause content that may be added to the contract.
We start with a transformer-based encoder-decoder architecture \cite{vaswani2017attention}, follow \cite{liu2019text} and initialize our encoder with {\sc ContractBERT}.
We then train the transformer decoder for generating clause content. As mentioned above, the inputs for the encoder comprise of a contract and a target clause type. 

We calculate the representations of all possible clauses belonging to the given type in the dataset using {\sc ContractBERT}, and their [CLS] token's embeddings are averaged, to obtain a target clause type representation \textit{trgt\_cls\_rep}.
This \textit{trgt\_cls\_rep} and the contract representation \textit{ct\_rep} are averaged to obtain the encoding of the given contract and target clause type, which is used as input to the decoder.
Note that since {\sc ContractBERT} is already pre-trained on the contracts, we do not need to train the encoder again for clause generation.
Given the average of the contract and target clause type representation as input, the decoder is trained to generate the appropriate clause belonging to the target type which might be relevant to the contract.
Note that our generation method provides a single clause as recommendation.
On the other hand, with retrieval-based methods, we can obtain multiple suggestions for a given clause type using similarity measures. 

\section{Experiments and Evaluation}

We evaluate three methods for clause type relevance prediction + clause recommendation:
{\bf (1)} Binary classification + clause generation, which is our proposed approach;
{\bf (2)} Collaborating filtering + similarity-based retrieval; and
{\bf (3)} Document similarity + similarity-based retrieval.

\noindent
\textbf{Collaborating filtering (CF) + similarity-based retrieval.}
Clause type relevance prediction can be seen as an {\it item-item based CF} task \cite{1167344} with contracts as users and clause types as items.
We construct a contract-clause type matrix, equivalent to the user-item matrix.
If contract $u$ contains clause type $i$, the cell $(u,i)$ gets the value $1$, otherwise $0$.
We then compute the similarity between all the clause type pairs ($i, j$), using an adjusted cosine similarity, given by, 
\begin{equation}
    sim(i,j) = \frac{\sum_u^U (r_{(u,i)} - \bar{r_u})(r_{(u,j)} - \bar{r_j})}{\sqrt{\sum_u^U r_{(u,i)}^2} \sqrt{\sum_u^U r_{(u,j)}^2}}
\end{equation}
We obtain the item similarity matrix using this cosine score, and
use it to predict if a target clause type $t$ is relevant to a given contract. 
We compute the score for $t$ using the weighted sum of the score of the other similar clause types, given by,

\begin{equation}
     score(u,t) = \frac{\sum_j^I sim(t,j)(r_{(u,j)}-\bar{r_j})}{\sum_j^I sim(t,j)} + \bar{r_t}
\end{equation}

If $t$ gets a high score and is not already present in the contract, it is recommended.
We experiment with multiple thresholds above which a clause type may be recommended. 
Given a clause library containing all possible clause types and their corresponding clauses, clause content recommendation can be seen as a {\it similarity-based retrieval} task.
For a given contract and a target clause type $t$, we use \textit{ct\_rep} and \textit{trgt\_cls\_rep}, and find cosine similarities with each of the clauses belonging to $t$ to find the most similar clauses that may be relevant to the given contract.
We do so by computing the similarity of either {\it (i)} \textit{ct\_rep} or {\it (ii)} (\textit{ct\_rep} + \textit{trgt\_cls\_rep})/2, with individual clause representations.

\noindent
\textbf{Document similarity + similarity-based retrieval.}
This is based on using similar documents to determine if a target clause type $t$ can be recommended for a given contract.
The hypothesis is that similar contracts tend to have similar clause types.
To find similar documents, we compute cosine similarities between the given contract's representations \textit{ct\_rep} with those of all the contracts in the (training) dataset to identify the top $k$ similar contracts.
If $t$ is present in any of the $k$ similar contracts and is not present in the given contract, it is recommended as a relevant clause type to be added to the contract.
We experiment with $k\in\{1,5\}$.
Similarity-based retrieval for clause content recommendation is same as above.


\noindent
\textbf{Metrics.} We evaluate the performance of clause type relevance prediction using precision, recall, accuracy and F1-score metrics, and that of the clause content recommendation using ROUGE \cite{lin-2004-rouge} score.

\noindent
\textbf{Data.}
\begin{table}[t]
\centering
\scalebox{0.6}{\begin{tabular}{p{2.1cm}ccccc}
    \toprule
    \textbf{\textsc{Clause Type}} & \textbf{\textsc{Method}} & \textbf{\textsc{Prec.}} & \textbf{\textsc{Rec.}} & \textbf{\textsc{Acc.}} & \textbf{\textsc{F1}} \\[0.5ex]
    \midrule
    
\textbf{Governing}
& CF-based  & 0.5889 & \textbf{0.8166} & 0.6243 & 0.6843 \\
\textbf{Laws} & Doc sim-based  &  \textbf{0.7882} & 0.6225 & \textbf{0.7276} & 0.6957 \\
& Binary classification & 0.6898 & 0.7535 & 0.7082 & \textbf{0.7203}  \\
\midrule

\textbf{Severability}
& CF-based  & 0.6396 & \textbf{0.9091} & 0.6987 & 0.7509 \\
& Doc sim-based & 0.7156 & 0.8182 & 0.7467 & 0.7635 \\
& Binary classification & \textbf{0.7654} & 0.8042 & \textbf{0.7790} &  \textbf{0.7843}  \\
\midrule

\textbf{Notices}
& CF-based  & 0.5533 & \textbf{0.8810} & 0.5885 & 0.6797 \\
& Doc sim-based  &  \textbf{0.7825} & 0.7257 & \textbf{0.7640} & \textbf{0.7530} \\
& Binary classification & 0.6850 & 0.7605 & 0.7079 & 0.7208  \\
\midrule

\textbf{Counterparts}
& CF-based  & 0.6133 & \textbf{0.8899} & 0.6657 & 0.7262 \\
& Doc sim-based  &  0.7156 & 0.8182 & 0.7467 & 0.7635 \\
& Binary classification & \textbf{0.7784} & 0.8259 & \textbf{0.7961} & \textbf{0.8014}  \\
\midrule

\textbf{Entire}
& CF-based  & 0.6197 & \textbf{0.8173} & 0.6591 & 0.7049 \\
\textbf{Agreements} & Doc sim-based & \textbf{0.9006} & 0.6623 & \textbf{0.7953} & 0.7633 \\
& Binary classification & 0.7480 & 0.8158 & 0.7713 & \textbf{0.7804}  \\
\bottomrule

\end{tabular}}
\captionsetup{justification=centering}
\caption{Clause type relevance prediction results.}
\label{results_clause_label_prediction}
\vspace{-0.15in}
\end{table}
We use the LEDGAR dataset introduced by \citet{tuggener-etal-2020-ledgar}.
It contains contracts from the U.S. Securities and Exchange Commission (SEC) filings website, 
and includes material contracts (Exhibit-10), such as shareholder agreements, employment agreements, etc.
The dataset contains 12,608 clause types and 846,274 clauses from around 60,000 contracts.
Further details on the dataset are provided in the appendix.

Since this dataset can not be used for our work readily, we preprocess it to create proxy datasets for clause type relevance prediction and clause recommendation tasks. 
For the former, for a target clause type $t$, we consider the labels {\it relevant} and {\it not relevant} for binary classification.
For {\it relevant} class, we obtain contracts that contain a clause corresponding to $t$, and remove this clause; given such a contract as input in which $t$ is not present, the classifier is trained to predict $t$ as relevant to be added to the contract.
For the {\it not relevant} class, we randomly sample an equal number of contracts that do not contain $t$ in them.
For recommendation, we use the contracts that contain $t$ ({\it i.e.}, the relevant class contracts); the inputs consist of the contract with the specific clause removed and $t$, and the output is the clause that is removed.
For both the tasks, we partition these proxy datasets into train ($60\%$), validation ($20\%$) and test ($20\%$) sets.
These ground truth labels (\{{\it relevant, not relevant}\} for the first task and the clause content for the second task) that we removed are used for evaluation.
The implementation details are provided in appendix.
\section{Results and Discussion}
Table \ref{results_clause_label_prediction} summarizes the results of the three methods (CF-based, document similarity-based and binary classification) for the clause type relevance prediction task.
For the tasks, we report results on the thresholds, $k$ and learning rate which gave best results on the validation set (the ablation results are reported in the appendix).

\begin{table}[t]
\centering
\scalebox{0.65}{\begin{tabular}{ccccc}
    \toprule
    \textbf{\textsc{Clause Type}} &
    \textbf{\textsc{Method}} &
    \textbf{\textsc{R-1}} &
    \textbf{\textsc{R-2}} &
    \textbf{\textsc{R-L}} \\
    \midrule
\textbf{Governing}
& Sim-based (w/o cls\_rep) & 0.441 & 0.213 & 0.327  \\
\textbf{Laws} & Sim-based (with cls\_rep) & 0.499 & 0.280 & 0.399 \\
& Generation-based & \textbf{0.567} & \textbf{0.395} & \textbf{0.506} \\

\midrule
\textbf{Severability}
& Sim-based (w/o cls\_rep) & 0.419 & 0.142 & 0.269  \\
& Sim-based (with cls\_rep) & 0.444 & 0.155 & 0.288 \\
& Generation-based & \textbf{0.521} & \textbf{0.264} & \textbf{0.432}\\
\midrule
\textbf{Notices}
& Sim-based (w/o cls\_rep) & 0.341 & 0.085 & 0.207  \\
& Sim-based (with cls\_rep) & 0.430 & 0.144 & 0.309 \\
& Generation-based & \textbf{0.514} & \textbf{0.271} & \textbf{0.422} \\
\midrule
\textbf{Counterparts} 
& Sim-based (w/o cls\_rep) & 0.466 & 0.214 & 0.406 \\
& Sim-based (with cls\_rep) & 0.530 & 0.279 & 0.474 \\
& Generation-based & \textbf{0.666} & \textbf{0.495} & \textbf{0.667} \\
\midrule
\textbf{Entire} 
& Sim-based (w/o cls\_rep) & 0.433 & 0.183 & 0.306 \\
\textbf{Agreements} & Sim-based (with cls\_rep) & 0.474 & 0.201 & 0.331 \\
& Generation-based & \textbf{0.535} & \textbf{0.312} & \textbf{0.485} \\

\midrule
\end{tabular}
}
\captionsetup{justification=centering}
\caption{Clause content recommendation results.}
\label{results_cls_content_prediction}
\vspace{-0.2in}
\end{table}
The CF-based method gives the best recall values for all the clause types, while the precision, accuracy and F1 scores are worse compared to the other two methods.
This method does not incorporate any contextual information of the contract clause content and relies only on the presence or absence of clause types to predict if a target type is relevant, thus resulting in high recall and low precision and F1 scores.
While the results of document similarity-based and classification methods are comparable, both have merits and demerits.
While the document similarity-based method is simpler and more extensible than  classification which requires training a new classifier for each new clause type, the former requires a large collection of possible contracts to obtain decent results (particularly the recall values), which may not be available always.
Further, the performance of document similarity method is dependent on $k$.
This can be seen in the lower recall values for the document similarity method compared to those of classification.
The storage costs associated with the contract collection can also become a bottleneck for the document similarity method.
Also, currently there is no way to rank the clauses in the similar contracts, and hence its recommendations cannot be scoped, while 
in classification, the probability scores can be used to rank the clause types for relevance.
On an average, the F1 scores for binary classification are highest compared to the other methods, while the accuracies are comparable with the document similarity method. 
Table \ref{results_cls_content_prediction} shows the results for clause content recommendation using similarity and generation-based methods.
For the sim-based method, we use the clause with the highest similarity to compute ROUGE.
The scores using only \textit{ct\_rep} are lower than those with \textit{trgt\_cls\_rep}.
This is expected as \textit{trgt\_cls\_rep} adds further information on the clause type for which the appropriate clauses are to be retrieved.
Finally, the generation-based method results in the best scores for clause recommendation, thus indicating the usefulness of our proposed approach for this task.
Some qualitative examples using both the methods are provided in appendix.

For clause content recommendation, we focused primarily on relevance (in terms of ROUGE). In general, retrieval-based frameworks, like the one we proposed, are mostly extractive in nature, and hence might be perceived as “safer” (or factual) to avoid any noise and vocabulary change in clauses that may be incorporated by generation methods, particularly in domains like legal. However, they can also end up retrieving clauses irrelevant to the contract context at times, as we note from their lower ROUGE scores, as retrieval is based on similarity heuristics which may not always capture relevance, while generation is trained to generate the specific missing clause in each contract. 

We also notice that generated clauses have lower linguistic variations in them, i.e., generated clauses belonging to one type often look alike. However, this is expected as most clauses look very similar with only a few linguistic and content variations. We believe because clauses have this repetitive nature, there is a large untapped opportunity to leverage NLP methods for legal text generation while accounting for the nuances and factuality in them, to build more accurate clause recommendation frameworks. We believe our work can provide a starting point for future works to build powerful models to capture the essence of legal text and aid in authoring them. In the future, we aim to focus on balancing the relevance and factuality of clauses recommended by our system. 

\section{Conclusions}

We addressed AI-assisted authoring of contracts via clause recommendation. We proposed \textsc{ClauseREC} pipeline to predict clause types relevant to a contract and generate appropriate content for them based on the contract content. The results we get on comparing our approach with similarity-based heuristics and traditional filtering-based techniques are promising, indicating the viability of AI solutions to automate tasks for legal domain. Efforts in generating long contracts are still in their infancy and we hope our work can pave way for more research in this area.

\bibliography{citations}

\begin{thebibliography}{10}
\expandafter\ifx\csname natexlab\endcsname\relax\def\natexlab#1{#1}\fi

\bibitem[{Bhagavatula et~al.(2018)Bhagavatula, Feldman, Power, and
  Ammar}]{bhagavatula-etal-2018-content}
Chandra Bhagavatula, Sergey Feldman, Russell Power, and Waleed Ammar. 2018.
\newblock \href {https://doi.org/10.18653/v1/N18-1022} {Content-based citation
  recommendation}.
\newblock In \emph{Proceedings of the 2018 Conference of the North {A}merican
  Chapter of the Association for Computational Linguistics: Human Language
  Technologies, Volume 1 (Long Papers)}, pages 238--251, New Orleans,
  Louisiana. Association for Computational Linguistics.

\bibitem[{Devlin et~al.(2019)Devlin, Chang, Lee, and
  Toutanova}]{devlin-etal-2019-bert}
Jacob Devlin, Ming-Wei Chang, Kenton Lee, and Kristina Toutanova. 2019.
\newblock \href {https://doi.org/10.18653/v1/N19-1423} {{BERT}: Pre-training of
  deep bidirectional transformers for language understanding}.
\newblock In \emph{Proceedings of the 2019 Conference of the North {A}merican
  Chapter of the Association for Computational Linguistics: Human Language
  Technologies, Volume 1 (Long and Short Papers)}, pages 4171--4186,
  Minneapolis, Minnesota. Association for Computational Linguistics.

\bibitem[{Lin(2004)}]{lin-2004-rouge}
Chin-Yew Lin. 2004.
\newblock \href {https://www.aclweb.org/anthology/W04-1013} {{ROUGE}: A package
  for automatic evaluation of summaries}.
\newblock In \emph{Text Summarization Branches Out}, pages 74--81, Barcelona,
  Spain. Association for Computational Linguistics.

\bibitem[{Linden et~al.(2003)Linden, Smith, and York}]{1167344}
G.~Linden, B.~Smith, and J.~York. 2003.
\newblock \href {https://doi.org/10.1109/MIC.2003.1167344} {Amazon.com
  recommendations: item-to-item collaborative filtering}.
\newblock \emph{IEEE Internet Computing}, 7(1):76--80.

\bibitem[{Liu and Lapata(2019)}]{liu2019text}
Yang Liu and Mirella Lapata. 2019.
\newblock \href {http://arxiv.org/abs/1908.08345} {Text summarization with
  pretrained encoders}.

\bibitem[{Malkiel et~al.(2020)Malkiel, Barkan, Caciularu, Razin, Katz, and
  Koenigstein}]{malkiel-etal-2020-recobert}
Itzik Malkiel, Oren Barkan, Avi Caciularu, Noam Razin, Ori Katz, and Noam
  Koenigstein. 2020.
\newblock \href {https://doi.org/10.18653/v1/2020.findings-emnlp.154}
  {{R}eco{BERT}: A catalog language model for text-based recommendations}.
\newblock In \emph{Findings of the Association for Computational Linguistics:
  EMNLP 2020}, pages 1704--1714, Online. Association for Computational
  Linguistics.

\bibitem[{Simonson et~al.(2019)Simonson, Broderick, and
  Herr}]{simonson-etal-2019-extent}
Dan Simonson, Daniel Broderick, and Jonathan Herr. 2019.
\newblock \href {https://doi.org/10.18653/v1/W19-2203} {The extent of
  repetition in contract language}.
\newblock In \emph{Proceedings of the Natural Legal Language Processing
  Workshop 2019}, pages 21--30, Minneapolis, Minnesota. Association for
  Computational Linguistics.

\bibitem[{Tuggener et~al.(2020)Tuggener, von D{\"a}niken, Peetz, and
  Cieliebak}]{tuggener-etal-2020-ledgar}
Don Tuggener, Pius von D{\"a}niken, Thomas Peetz, and Mark Cieliebak. 2020.
\newblock \href {https://www.aclweb.org/anthology/2020.lrec-1.155} {{LEDGAR}: A
  large-scale multi-label corpus for text classification of legal provisions in
  contracts}.
\newblock In \emph{Proceedings of the 12th Language Resources and Evaluation
  Conference}, pages 1235--1241, Marseille, France. European Language Resources
  Association.

\bibitem[{Vaswani et~al.(2017)Vaswani, Shazeer, Parmar, Uszkoreit, Jones,
  Gomez, Kaiser, and Polosukhin}]{vaswani2017attention}
Ashish Vaswani, Noam Shazeer, Niki Parmar, Jakob Uszkoreit, Llion Jones,
  Aidan~N. Gomez, Lukasz Kaiser, and Illia Polosukhin. 2017.
\newblock \href {http://arxiv.org/abs/1706.03762} {Attention is all you need}.

\bibitem[{Wang and Fu(2020)}]{wang-fu-2020-item}
Tian Wang and Yuyangzi Fu. 2020.
\newblock \href {https://doi.org/10.18653/v1/2020.ecnlp-1.8} {Item-based
  collaborative filtering with {BERT}}.
\newblock In \emph{Proceedings of The 3rd Workshop on e-Commerce and NLP},
  pages 54--58, Seattle, WA, USA. Association for Computational Linguistics.

\end{thebibliography}
\bibliographystyle{acl_natbib}

\appendix
\section*{Appendix}
\label{sec:appendix}
\pdfoutput=1







%
%

\title{\textsc{ClauseRec:} A Clause Recommendation Framework for AI-aided Contract Authoring}


\author{First Author \\
  Affiliation / Address line 1 \\
  Affiliation / Address line 2 \\
  Affiliation / Address line 3 \\
  \texttt{email@domain} \\\And
  Second Author \\
  Affiliation / Address line 1 \\
  Affiliation / Address line 2 \\
  Affiliation / Address line 3 \\
  \texttt{email@domain} \\}
  
\def\vinay#1{\textcolor{brown}{[Vinay: #1] }}
\def\aparna#1{\textcolor{red}{[Aparna: #1] }}
\def\bvs#1{\textcolor{blue}{[Balaji: #1] }}


\section{Data}

\begin{figure}[t]
    \centering
    \includegraphics[width=0.9\linewidth]{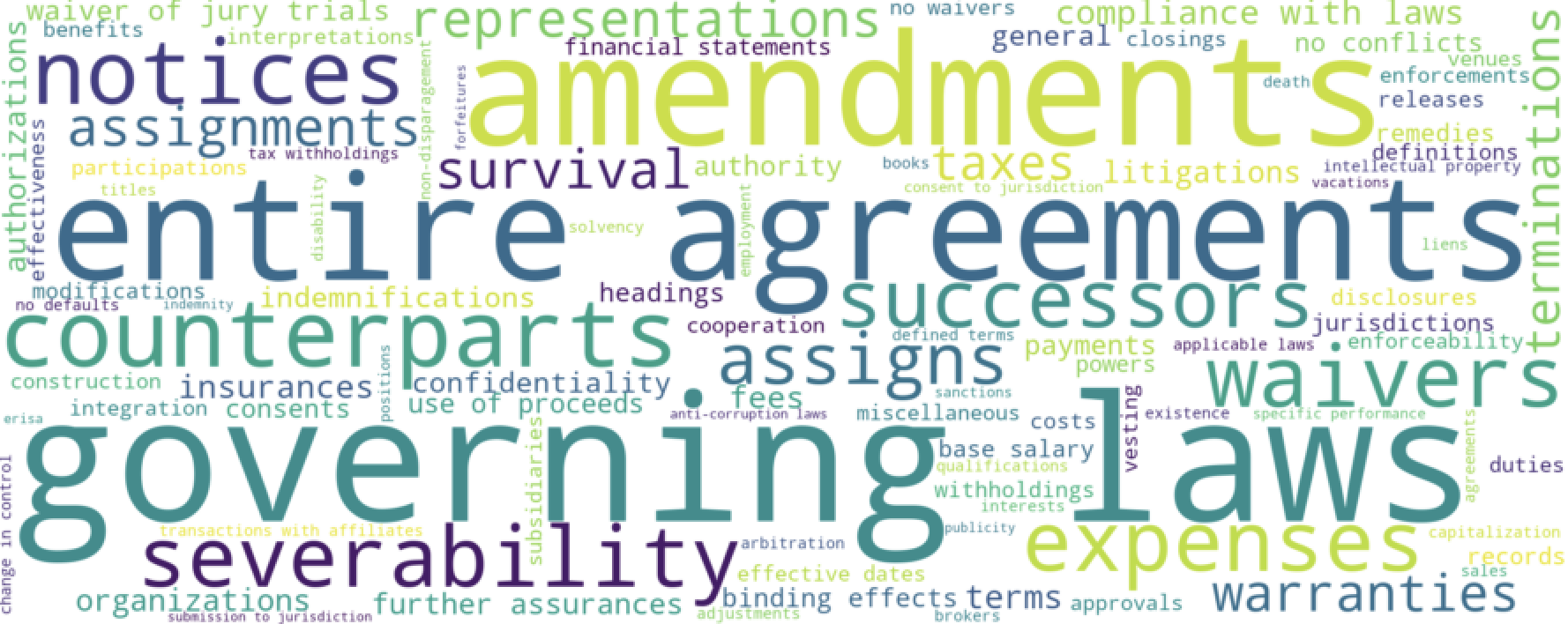}
    \captionsetup{justification=centering}
    \caption{Some clause types in the LEDGAR dataset.}
    \label{fig:clause_word_cloud}
    \vspace{-0.1in}
\end{figure}

Figure \ref{fig:clause_word_cloud} shows some of the clause types present in the LEDGAR dataset.

\section{Implementation Details}
To train {\sc ContractBERT}, we crawl and use a larger collection of 250k contracts and train it till the losses converge.
\subsection{Binary Classifiers}

We use a small 7-layered fully connected neural network with ReLU activation and dropout of $0.3$ as binary classifiers. The input is $768$ dimensional contract representation and output is a probability score between $[0,1]$. We use a batch-size of 64 and train them for 5000 epochs. We experiment with 4 learning rates: ${[1e-5,5e-6,1e-6,5e-7]}$. \textit{Adam} optimizer is used with Binary Cross Entropy Loss as criterion. The model with highest accuracy on validation set is stored and the results are reported on a held out test set. The training takes around 150 minutes for each clause type. 

For document similarity method, we experimented with $k = [1,5]$ and for CF based method, we evaluated F-scores and accuracies for different threshold values and report the best results we got. Table \ref{tab:implementation_cls_label} summarizes the corresponding $k$ values, $thresholds$ and $learning_rates$ corresponding to the best results.

\subsection{Transformer Decoder}

The clause text is pre processed by removing punctuation, single letter words, and multiple spaces then using nltk's word tokenizer \footnote{https://www.nltk.org/} to tokenize the clause text. We keep the maximum generation length to be 400 including <SOS> and <EOS> tokens. All the clauses with more than 398 tokens are discarded. The vocabulary is 7185 token long which is the output dimension. We use 3 decoder layers. The hidden dimension is 768 i.e., the length of input embedding. A dropout of $0.1$ is used. A constant learning rate of $1e-05$ is used with a batch size of 16 and the training takes place for 300 epochs. Validation split of $0.2$ is used. The results are reported on a held out test. 

\begin{table}[]
    \centering
    \scalebox{0.75}{\begin{tabular}{||c|c|c|c||}
    \hline
     \textbf{Clause Label} & \textbf{k-value} & \textbf{threshold} & \textbf{learning\_rate} \\
         
         \hline \hline
         \textbf{Governing Law} & \textbf{$1$} & \textbf{$0.27$} & \textbf{$5e-07$} \\
         \hline
         \textbf{Counterparts} & \textbf{$2$} & \textbf{$0.18$} & \textbf{$1e-06$} \\
         \hline
         \textbf{Notices} & \textbf{$2$} & \textbf{$0.15$} & \textbf{$5e-06$} \\
         \hline
         \textbf{Entire Agreements} & \textbf{$1$} & \textbf{$0.20$} & \textbf{$1e-05$} \\
         \hline
         \textbf{Severability} & \textbf{$3$} & \textbf{$0.13$} & \textbf{$1e-06$} \\ [1ex]
         \hline
    \end{tabular}}
    \caption{Implementation Details for Clause Type Prediction}
    \label{tab:implementation_cls_label}
\end{table}

\section{Qualitative Results}

Table \ref{tab:qualitative_results_rem} shows the qualitative results for a few clause types comparing the similarity-based retrieval with generation-based methods. The ROUGE-1 F-scores are mentioned in the brackets to compare the results quantitatively as well. 
\begin{table*}[t]
\scalebox{0.65}{\begin{tabular}{  l c p{18cm} }
    \toprule
    \textbf{\textsc{Clause Type}}      
    & \textbf{}   
    & \textbf{\textsc{Clause}}\\
    \midrule
    
    \multirow{3}{*}{\textbf{Governing Laws}} 
    & 
    Original & This agreement and the obligations of the parties here under shall be governed by and construed and enforced in accordance with the substantive and procedural laws of the state of delaware without regard to rules on choice of law. \\
    &
    Sim-based & This agreement shall be governed by and construed in accordance with the laws of the state of illinois without giving effect to the principles of conflicts of law rules the parties unconditionally and irrevocably consent to the exclusive jurisdiction of the courts located in the state of illinois and waive any objection with respect thereto for the purpose of any action suit or proceeding arising out of or relating to this agreement or the transactions contemplated hereby. \textbf{(R1: 0.456)} \\
    &
    Generated & This agreement shall be governed by and construed in accordance with the laws of the state of delaware without regard to the conflicts of law principles thereof. \textbf{(R1: 0.718)} \\
    \midrule
    
    \multirow{3}{*}{\textbf{Notices}} 
    &
    Original & Any notices required or permitted to be given under this agreement shall be sufficient if in writing and if personally delivered or when sent by first class certified or registered mail postage prepaid return receipt requested in the case of the executive to his residence address as set forth in the books and records of the company and in the case of the company to the address of its principal place of business to such person or at such other address with respect to each party as such party shall notify the other in writing. \\
    & 
    Sim-based & Any notice required or permitted by this agreement shall be in writing and shall be delivered as follows with notice deemed given as indicated by personal delivery when delivered personally ii by overnight courier upon written verification of receipt iii by telecopy or facsimile transmission upon acknowledgment of receipt of electronic transmission or iv by certified or registered mail return receipt requested upon verification of receipt notice shall be sent to executive at the address listed on the company personnel records and to the company at its principal place of business or such other address as either party may specify in writing. \textbf{(R1: 0.588)} \\  
    &
    Generated & Any notice required or permitted to be given under this agreement shall be sufficient if in writing and if sent by registered or certified mail return receipt requested to executive at the last address the executive has filed in writing with the company or in the case of the company to its main offices attention of the board. \textbf{(R1: 0.640)} \\
    \midrule

    \multirow{3}{*}{\textbf{Severability}} 
    &
    Original & If any term provision covenant or restriction of this agreement is held by court of competent jurisdiction to be invalid illegal void or unenforceable the remainder of the terms provisions covenants and restrictions set forth herein shall remain in full force and effect and shall in no way be affected impaired or invalidated and the parties hereto shall use their commercially reasonable efforts to find and employ an alternative means to achieve the same or substantially the same result as that contemplated bit/s]y such term provision covenant or restriction it is hereby stipulated and declared to be the intention of the parties that they would have executed the remaining terms provisions covenants and restrictions without including any of such that may be hereafter declared invalid illegal void or unenforceable. \\
    &
    Sim-based & Each party hereto agrees that should any court or other competent authority hold any provision of this agreement or part of this agreement to be null void or unenforceablen the or order any party to take any action inconsistent herewith or not to take an action consistent with the terms of or required by this agreement the validity legality and  or enforceability of the remaining provisions and obligations contained or set forth in this agreement shall not in any way be affected or impaired unless the foregoing inconsistent action or the failure to take an action constitutes material breach of this agreement or makes this agreement impossible to perform in which case this agreement shall terminate. \textbf{(R1: 0.396)} \\
    & 
    Generated & If any provision of this agreement is held to be invalid or unenforceable under present or future laws effective during the term hereof the term hereof the term or provisiagenton shall be fully severable and this agreement shall be construed and enforced as if such illegal invalid or unenforceable term or provision had never comprised part hereohim hf and the remaining terms hereof shall remain in full force and effect and shall not be affected by the illegal invalid or unenforceable term or provision or by the parties hereto shall negotiate in good faith substitute valid and enforceable provision that most nearly reflects the original intent of the parties and all other provisions hereof shall remain in full force and effect and shall be enforceable. \textbf{(R1: 0.504)} \\ 
    \midrule
    
    \multirow{3}{*}{\textbf{Counterparts}} 
    &
    Original & This agreement may be executed in one or more counterparts each of which will constitute an original but all of which when taken together will constitute but one contract and will become effective as provided in section hereof delivery of an executed counterpart to this agreement by facsimile or other electronic transmission will be as effective as delivery of manually signed original. \\
    &
    Sim-based & This agreement may be executed by the parties hereto in separate counterparts each of which when so executed and delivered shall be an original but all such counterparts shall together constitute one and the same agreement and all signatures need not appear on any one counterpart executed copies of the signature pages of this agreement sent by facsimile or transmitted electronically in portable document format or any similar format shall be treated as originals fully binding and with full legal force and effect and the parties waive any rights they may have to object to such treatment. \textbf{(R1: 0.427)} \\
    &
    Generated & This agreement may be executed in any number of counterparts and by different parties hereto in separate counterparts each of which when so executed shall be deemed to be an original and all of which taken together shall constitute one and the same agreement delivery of an executed counterpart of signature page to this agreement by facsimile or other electronic imaging means shall be effective as delivery of manually executed counterpart of this agreement. \textbf{(R1: 0.603)} \\
    \midrule

    \multirow{3}{*}{\textbf{Entire}} 
    &
    Original & This agreement constitutes the entire agreement of the signing parties with respect to the subject matter hereof and supersedes all other understandings oral or written with respect to the subject matter hereof there are no oral or implied obligations of the control agent or the other lenders to any third party in connection with this agreement. \\
    \textbf{Agreements} & Sim-based & This agreement consisting of sections through with schedules and the technology license agreement which is expressly incorporated by reference herein constitutes the entire understanding between the parties concerning the subject matter hereof and supersedes all prior discussions agreements and representations whether oral or written this agreement may be amended altered or modified only by an instrument in writing duly executed by the authorized representations of both parties. \textbf{(R1: 0.435)} \\
    &
    Generated & This agreement and the other transaction documents constitute the entire agreement among the parties hereto with respect to the subject matter hereof and thereof and supersede all other prior agreements and understandings both written and oral among the parties or any of them with respect to the subject matter hereof. \textbf{(R1: 0.626)} \\[1ex]

    \bottomrule
    \end{tabular}}
    \captionsetup{justification=centering}
    \caption{Qualitative comparison of retrieved and generated clauses}
    \label{tab:qualitative_results_rem}
\end{table*}







\end{document}